\documentclass{imag-ms-template}

\title{Structural Connectome Harmonization Using Deep Learning:   \\           The Strength of Graph Neural Networks} 

\author{Jagruti Patel,$^{1\ast}$ Thomas A. W. Bolton,$^{1}$ Mikkel Schöttner,$^{1}$ Anjali Tarun,$^{1}$ \\Sebastien Tourbier,$^{1}$ Yasser Alemán-Gómez,$^{1}$ Jonas Richiardi,$^{1}$ Patric Hagmann$^{1}$\\
{\small $^{1}$Department of Radiology, Lausanne University Hospital and University of Lausanne (CHUV-UNIL),}\\
{\small Lausanne, Switzerland}\\
{\small $^\ast$Correspondence:  jagruti90@gmail.com}
}

\begin{document} 

\maketitle


\begin{abstract}
Small sample sizes in neuroimaging in general and in structural connectome (SC) studies in particular limit the development of reliable biomarkers for neurological and psychiatric disorders—such as Alzheimer’s disease and schizophrenia—by reducing statistical power, reliability, and generalizability. Large-scale multi-site studies have exist, but such studies have acquisition-related biases due to scanner heterogeneity, compromising imaging consistency and downstream analyses. While existing SC harmonization methods—such as linear regression, ComBat, and deep learning (DL) techniques—mitigate acquisition-related biases, they often rely on detailed metadata, traveling subjects, or overlook the graph-topology of SCs. To address these limitations, we propose a site-conditioned deep harmonization framework that harmonizes SCs across diverse acquisition sites without requiring metadata or traveling subjects that we test in a simulated scenario based on the Human Connectome Dataset. Within this framework, we benchmark three deep architectures—a fully connected autoencoder, a convolutional autoencoder, and a graph convolutional autoencoder—against a top-performing linear regression baseline. While non-graph models excel in edge-weight prediction and edge existence detection, the graph autoencoder demonstrates superior preservation of topological structure and subject-level individuality, as reflected by graph metrics and fingerprinting accuracy, respectively. Although the linear baseline achieves the highest numerical performance by explicitly modeling acquisition parameters, it lacks applicability to real-world multi-site use cases as detailed acquisition metadata is often unavailable. Our results highlight the critical role of model architecture in SC harmonization performance and demonstrate that graph-based approaches are particularly well-suited for structure-aware, domain-generalizable SC harmonization in large-scale multi-site SC studies.
\end{abstract}

\section{Introduction}

Modern neuroimaging techniques, such as Magnetic Resonance Imaging (MRI), have revolutionized neuroscience research by enabling the delineation of brain regions and the computation of their connections to generate detailed brain maps. Computed in physical space, structural connectomes (SCs) represent these connections and offer insights into the brain’s structural network organization \citep{hagmann2007mapping, hagmann2005diffusion, sporns2005human}. SCs have been instrumental in studying a variety of brain-related disorders, including Alzheimer’s disease, schizophrenia, motor neuron disease, multiple sclerosis, and epilepsy \citep{wu2021clinical, basaia2020structural, griffa2013structural}. Despite this promise, research leveraging SCs faces a major barrier: small sample sizes in most studies limit the statistical power, generalizability, and clinical applicability of findings \citep{pinto2020harmonization, orban2018multisite}. Aggregating data from multiple sites has become a potential solution, allowing for the creation of larger, more diverse multisite datasets \citep{pearlson2009multisite}. However, SCs from multiple sites  are susceptible to scanner-related biases arising from differences in diffusion encoding schemes, acquisition parameters, such as spatial/angular resolution, and strength and direction of b-values of Diffusion-weighted imaging (DWI) \citep{patel2024modeling, borrelli2022structural, sotiropoulos2019building, caiazzo2018structural, gigandet2013connectome}. For example, higher angular resolution methods like Diffusion spectrum imaging (DSI) recover fiber connections missed by Diffusion tensor imaging (DTI), resulting in denser-looking brain maps for the same individual \citep{gigandet2013connectome}. Such inconsistencies across acquisition settings complicate downstream analyses and reduce the sensitivity of SC-based findings.

To address these issues, several harmonization approaches have been proposed. One strategy is correcting structural-weighted MRI and/or DWI at the source before generating SCs. Diffusion data can be harmonized at the signal-level using rotation-invariant spherical harmonic features \citep{karayumak2019retrospective} or at the imaging-level with deep learning (DL) methods \citep{ning2020cross}. DL techniques such as CycleGAN \citep{roca2025iguane}, StyleGAN \citep{liu2023style}, and autoencoders \citep{gebre2023cross, cackowski2023imunity, fatania2022harmonisation} have been employed for harmonizing multi-site structural-weighted MRI. However, applying these methods in the SC analysis pipeline may propagate errors introduced during signal/image-space harmonization, amplifying inaccuracies in computing fiber ODFs during tractography and ultimately in the final SCs \citep{kurokawa2021cross}.

This motivates to harmonize SCs directly at the matrix-level, using either linear or DL techniques. Linear models such as ComBat have been effective in removing site effects \citep{onicas2022multisite}, while generalized linear models assuming gamma-distributed connectivity have been proposed as alternatives to ComBat’s normal distribution assumption \citep{shen2024harmonization}. Alternatively, traveling subjects—individuals scanned at multiple acquisition sites—enable SC correction based on overlapping biological profiles \citep{kurokawa2021cross}. \citep{patel2024modeling} introduced an effective linear regression method to mitigate SC bias by explicitly modeling acquisition parameters as regression covariates. While effective, these methods pose practical challenges, such as requiring labeled metadata (\textit{e.g.}, acquisition parameters, scanner vendor or version) or relying on traveling subjects, which may be infeasible for large-scale SC-based studies. SC-based fingerprinting \citep{ciarrusta2022developing}, requiring explicit consideration of all biological covariates for each individual, is also impractical. Moreover, modeling scanner-related variations across multiple sites is cumbersome, especially across vendors or scanner versions.

Recent advancements in DL have introduced non-linear harmonization methods, leveraging autoencoders for extracting latent, site-invariant features. For instance, \citep{newlin2024learning} proposed a fully connected 1D autoencoder that vectorized SCs to extract site-invariant features and reconstructed harmonized SCs by conditioning the decoder on site-specific variables. While effective at removing site bias, these models treated connectomes as unstructured feature vectors, disregarding the topological and region-specific properties inherent to the brain network. 2D Convolutional Neural Networks (2D-CNNs), while extensively applied in harmonizing multi-site MRI data, remain underexplored for SC harmonization. By employing local, reusable, and translation-invariant convolutional filters, CNNs excel at capturing local patterns and spatial relations, making them a promising candidate for SC harmonization. However, 2D-CNNs are inherently constrained by their assumption of spatial locality and struggle to capture the global SC structure, where meaningful relationships often exist between non-adjacent entries, reflecting the non-local organization of brain networks.

Graph Neural Networks (GNNs) have emerged as a powerful tool for analyzing graph-structured SC data \citep{ktena2017distance}. By leveraging message passing and aggregation, GNNs effectively capture both global and local topological patterns while preserving node identities and edge relationships. They have been successfully applied to various SC-based tasks, including brain age prediction, disease classification, and graph-to-graph translation \citep{gao2023brain, he2023predicting, hong2023structural, zhang2022predicting, yao2021mutual}. Towards SC harmonization, \citep{zhang2021disentangled} have attempted to integrate heterogeneous connectomes from multiple views into a shared unified latent space. However, the method relies on multiple encoders and decoders for each view, presenting scalability issues as the number of acquisition sites grows. Also, the method requires paired connectomes across views—a prerequisite that is rarely available in practice. Despite the strengths of GNNs, their application for SC harmonization remains underexplored yet promising direction.

In this paper, our goal is two-fold:
\begin{enumerate}
    \item To develop a scalable and data-efficient deep framework for SC harmonization, and
    \item To systematically evaluate the strengths and limitations of various linear and non-linear deep models within the framework — with a particular focus on the representational benefits offered by GNNs.
\end{enumerate}

To achieve this, we propose a generalized deep framework for harmonizing SC — accurately predicting SCs from any source site to any target site. 
Our major contributions include:
\begin{enumerate}
    \item A site-conditioned deep harmonization framework with a single-encoder-single-decoder design that scales seamlessly across multiple sites learning site-invariant representations from independently acquired SCs.
    \item The framework generalizes across diverse acquisition parameters without requiring metadata, traveling subjects, or site-specific encoders. 
    \item The framework adapts incorporates data augmentation techniques, inspired by image-based approaches, to mitigate SC dataset size limitations, enhancing robustness and generalizability. 
    \item  We benchmark three deep architectures within the framework—fully connected 1D autoencoder, 2D convolutional autoencoder, and graph autoencoder—trained adversarially with a site-classifier to enforce invariance and a site-conditioned decoder for reconstructing SCs in a target domain.
    \item Comprehensive evaluations across key metrics, including edge-level accuracy, network topology preservation, subject identifiability and demographic prediction (age and gender) and a thorough benchmarking against a top-performing linear regression baseline  \citep{patel2024modeling}.
\end{enumerate}

We conducted this study within a simulated framework using resampled data derived from the high-quality Human Connectome Project Young Adult (HCP-YA) dataset \citep{van2012human, glasser2013minimal, van2013wu, sotiropoulos2013advances}, leveraging the availability of ground truth to rigorously validate our findings.

\section{Materials and Methods}


\subsection{Dataset}
Minimally preprocessed T1-weighted (T1-W) and DW-MRI data were downloaded for 1,065 subjects from the HCP-YA dataset. Additionally, similar data were obtained for a subset of 44 subjects who underwent a second scanning session within a period of 18 to 343 days. To improve computational efficiency, T1-W images were downsampled from 0.7 mm to 1.25 mm isotropic resolution. DWI data was resampled to simulate variability in acquisition parameters based on b-value and spatial resolution. The original DWI data, acquired with multiple b-value shells (0, 1000, 2000, 3000 s/$\text{mm}^2$)) — with approximately equal numbers of gradient directions for each non-zero b-value shell and 18 b0 images — and 1.25 mm isotropic resolution, was first downsampled to 2.3 mm using trilinear interpolation. For both the original and downsampled images, DWI data corresponding to b-value shells of 1000 $\pm$ 20 and 3000 $\pm$ 20, along with b-values of 0–15, were extracted. This resampling effectively mimicked a scenario where each subject was scanned under four distinct acquisition parameters (\textbf{Table \ref{tab:site_params}}).

Structural connectivity matrices were computed using Connectome Mapper 3 (version 3.0.0-rc4) \citep{tourbier2022connectome, tourbier2020connectomicslab} from the T1-W images and DWI data. T1-W images were first parcellated into brain regions using Freesurfer recon-all (version 6.0.1), based on the Lausanne 2018 scale-3 atlas \citep{cammoun2012mapping, iglesias2015computational, iglesias2015bayesian, najdenovska2018vivo, desikan2006automated}, which divides the brain into 274 regions \citep{patel2024modeling}. The DWI data, already aligned with the T1-W images, was then processed with MRtrix3 \citep{tournier2012mrtrix} to generate tractograms. Specifically, Fiber Orientation Distribution Functions (fODFs) were estimated using constrained spherical deconvolution (CSD) \citep{tournier2007robust} of order 8, and then, deterministic tractography was performed using white matter seeding to generate 10 million streamlines. These tractograms were then combined with the parcellated T1-W images to construct subject-specific SCs, where each edge in the SC represented the number of streamlines connecting a pair of regions, and diagonal self-connections were set to zero \citep{borrelli2022structural}.

\begin{figure}[t]
\centerline{\includegraphics[width=0.95\textwidth]{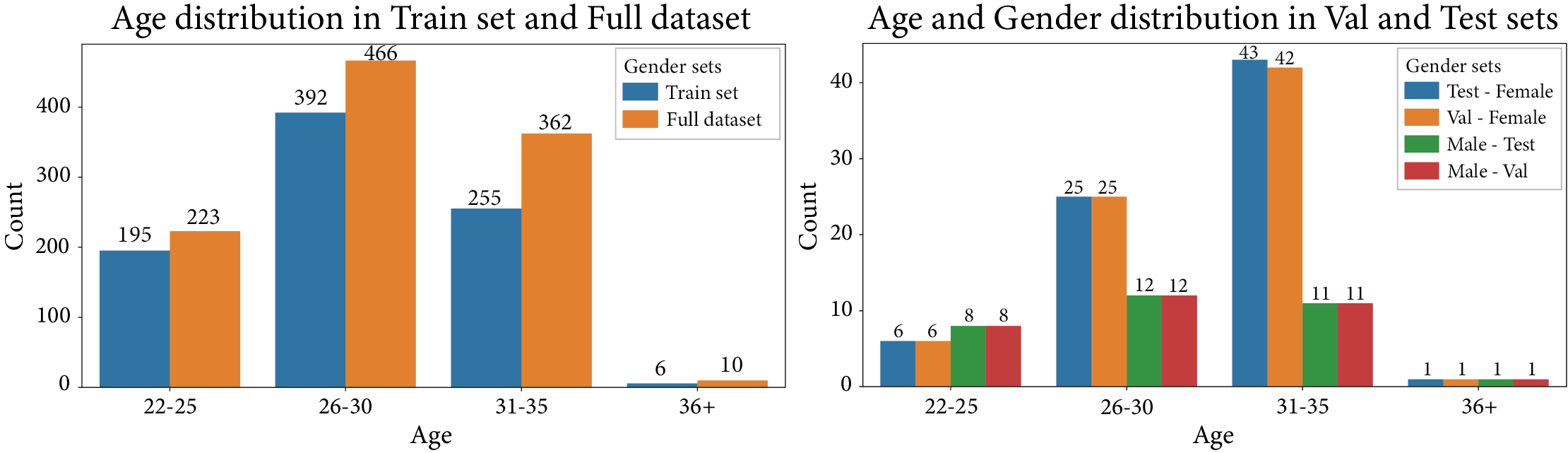}}
\caption{Age and gender distribution across different data splits.}
\label{fig:ageGenRange}
\end{figure}

\subsubsection{Dataset splitting}
A total of 1,061 subjects were selected based on the availability of brain connectivity matrices at scale 3 across all four acquisition parameter combinations. Among the 44 test-retest subjects, 2 were excluded—one due to preprocessing issues, and another due to inter-scan differences comparable to inter-scan inter-subject difference \citep{patel2024modeling}. The dataset was systematically partitioned into training, validation, and test sets in an 80:10:10 ratio, ensuring that subjects with shared family IDs were assigned to the same group to eliminate the risk of data leakage \citep{ooi2022comparison}. Additionally, test-retest subjects and their related subjects were excluded from the training and validation sets. This split ultimately resulted in 848 subjects for training, 106 for validation, and 107 for testing, and each subject included SCs across four acquisition parameter combinations. The training set was balanced for gender, with subjects filtered to ensure an age distribution closely aligned with the overall dataset, shown in \textbf{Figure \ref{fig:ageGenRange}}. Similarly, age and gender distributions for the validation and test sets were matched to preserve consistency across groups, shown in \textbf{Figure \ref{fig:ageGenRange}}. 

\begin{figure}[t]
\centerline{\includegraphics[width=0.95\textwidth]{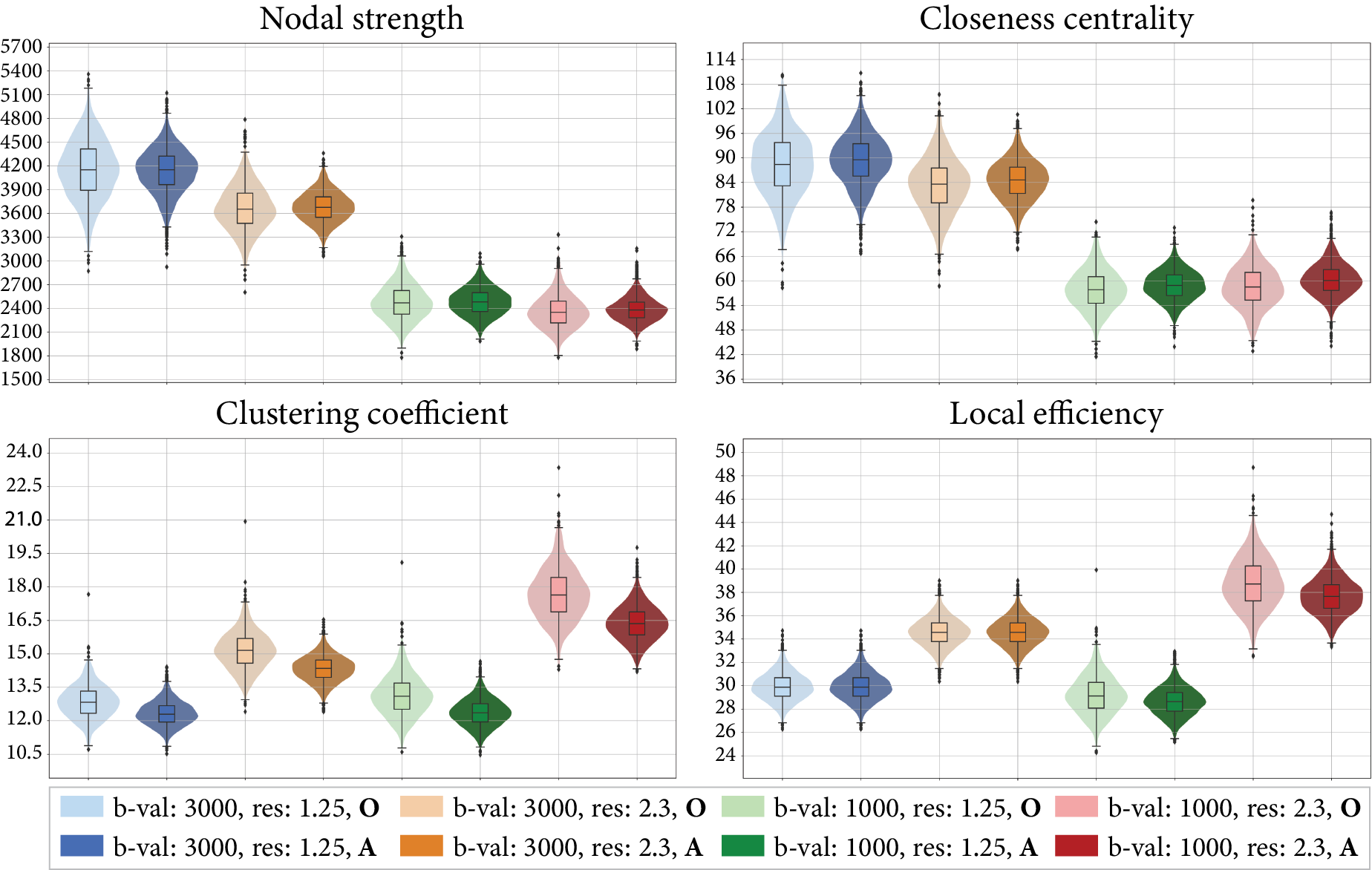}}
\caption{Comparison of different graph metrics between original (\textbf{O}) and augmented (\textbf{A}) structural connectivity matrices across acquisition parameter combinations in the training set.}
\label{fig:augmentation}
\end{figure}

\subsubsection{Training and augmentation}
%
For training without using traveling-subject data, each of the 848 subjects was randomly assigned a SC generated using one of the four acquisition parameter combinations, ensuring equal distribution across the combinations in the training set, i.e. imaging site. Subsequently, the training set for each acquisition parameter combination was expanded using a tailored data augmentation strategy, inspired by Mixup \citep{zhang2017mixup}. For each acquisition parameter combination, pairs of subjects were randomly selected and their connectivity mixed in the following way. The upper triangular part of their structural connectivity matrices was vectorized (excluding the diagonal). A random binary mask of the vector length was applied to generate an augmented vector: fibers from the first subject were selected where the mask value was one, while fibers from the second subject were selected where the mask value was zero. This augmented vector was reconstructed into a symmetric structural connectivity matrix to preserve structural integrity. Using this approach, 4,000 augmented matrices were generated per acquisition parameter combination, totaling 16,848 SCs for training.
To validate the consistency of the augmented SCs with the original ones, key graph metrics—including nodal strength, closeness centrality, clustering coefficient, and local efficiency—were computed and compared \citep{rubinov2010complex, patel2024modeling}. \textbf{Figure \ref{fig:augmentation}} displays these metrics for each acquisition parameter combination, comparing the original 848 SCs to the augmented 4,000 SCs. 
The original and augmented SC distributions align closely across all metrics and acquisition parameter combinations, except for the clustering coefficient. As noted in \citep{patel2024modeling}, the clustering coefficient is sensitive to the highest edge weight (\textit{i.e.}, fiber count). Random selection of lower maximum edge weights during augmentation likely caused the slight reduction in the mean clustering coefficient in the augmented data.

\subsubsection{Evaluation}
We assessed the model using both the validation and test sets, which contained SCs from all four acquisition parameter combinations. All SCs were harmonized to the highest b-value of 3000 and the highest isotropic spatial resolution of 1.25 mm. The evaluation also included SCs from the 42 test-retest subjects, at the highest b-value and spatial resolution.

\subsection{SC Harmonization Methods}
This section describes methods for harmonizing SCs, including linear regression (LR) and various DL models: 1D fully connected autoencoder (FAE), 2D convolutional autoencoder (CAE), and graph autoencoder (GAE). Despite differences in their building blocks, all DL models follow a standardized SC harmonization framework, as illustrated in \textbf{Figure \ref{fig:dl_framework}}. The training procedure for these models is consistent, with minor variations detailed in the respective sections.

\subsubsection{\textbf{\textit{Notations:}}}
In this study, the dataset $\mathcal{D}$ consists of $|\mathcal{D}|$ SCs. 
The site-label for the $i^{th}$ subject is denoted as $y_i^c$, where $\mathcal{C}$ is the number of sites or acquisition parameter combinations and $c \in \mathcal{C}$ is the site-variable. The corresponding structural connectivity matrix is denoted as $\mathbf{S}_i^c \in \mathbb{R}^{N \times N}$, where $N$ is the number of regions of interest (RoIs). The site details with respective acquisition parameters are presented in \textbf{Table~\ref{tab:site_params}}.

\subsubsection{\textit{\textbf{Linear Regression for SC Harmonization:}}}
LR is formulated as outlined in \cite{patel2024modeling}, where each connection $s_i^c \in \mathbf{S}_i^c$ in the upper triangle of a structural connectivity matrix is independently modeled, assuming independence among the connections, as a weighted linear combination of the acquisition parameters and their interactions.
\[
s_i^c = \beta_{0} + \beta_{1} * X_r^c + \beta_{2} * X_b^c + \beta_{3} * X_r^c * X_b^c + \varepsilon_i^c 
\]
where ${[\beta_0, \beta_1, \beta_2, \beta_3]}^\top$ is the weight vector to be estimated per connection; $X_r^c$ and $X_b^c$ denote the respective \textit{resolution} and \textit{b-value}; and error $\varepsilon_i^c$ is random, following a Gaussian distribution with zero-mean. Notably, the method does not explicitly incorporate subject-specific covariates, which allows to harmonize unseen data.
Upon training, LR yields a vector of ordinary least squares estimates, ${[\hat{\beta}_0, \hat{\beta}_1, \hat{\beta}_2, \hat{\beta}_3]}^\top$. During inference, the structural connection value $s_i^k \in \mathbf{S}_i^k$ for an unseen $k^\text{th}$ site is estimated as, 
\[
s_i^k = s_i^c + \hat{\beta}_1 * (X_r^k-X_r^c) + \hat{\beta}_2 * (X_b^k-X_b^c) + \hat{\beta}_3 * (X_r^k * X_b^k - X_r^c * X_b^c)
\]
To ensure biological plausibility, negative estimated values were first set to zero, followed by rounding all values to the nearest integer to accommodate the discrete nature of fiber counts.

\begin{figure}
\centerline{\includegraphics[width=0.95\textwidth]{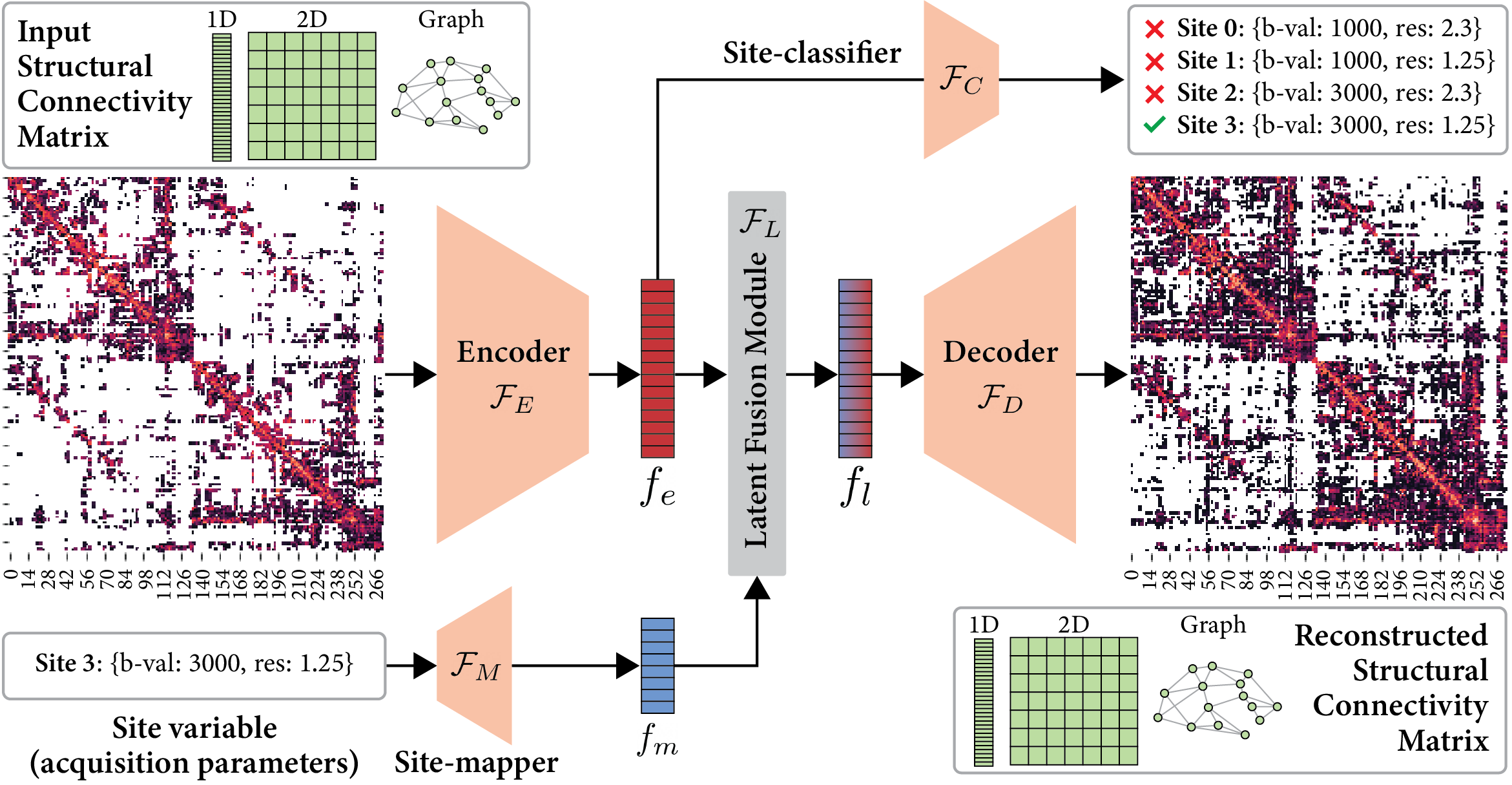}}
\caption{Overview of our deep SC harmonization framework, featuring an encoder-decoder architecture, site-mapper, site-classifier, and latent-fusion module. The framework supports encoder-decoder architectures handling 1D, 2D, or graph input formats of structural connectivity matrix.}
\label{fig:dl_framework}
\vspace{-4pt}
\end{figure}

\begin{table}[t]
    \centering
    \renewcommand{\arraystretch}{1.0}
    \begin{tabular}{c|c|c}
        \toprule
        \textbf{B-value ($s/{mm}^2$}) & \textbf{Spatial resolution (${mm}^3$)} & \textbf{Site condition variable} \\ 
        \midrule
        1000 & 2.3 & 0 \\
        1000 & 1.25 & 1 \\
        3000 & 2.3 & 2 \\
        3000 & 1.25 & 3 \\
        \bottomrule
    \end{tabular}
    \caption{Diffusion acquisition parameters across sites.}
    \label{tab:site_params}
\end{table}

\subsubsection{Deep Learning-based SC Harmonization}
In this section, we present our proposed unified SC harmonization framework, consisting of five core modules: \textbf{encoder} ($\mathcal{F}_E$), \textbf{decoder} ($\mathcal{F}_D$), \textbf{site-classifier} ($\mathcal{F}_C$), \textbf{site-mapper} ($\mathcal{F}_M$), and \textbf{latent-fusion} ($\mathcal{F}_L$) (\textbf{Figure \ref{fig:dl_framework}}). While $\mathcal{F}_E$ and $\mathcal{F}_D$ are implemented using different deep learning models such as FAE, CAE, or GAE, $\mathcal{F}_C$ and $\mathcal{F}_M$ adopt consistent architectures across all methods.
\vspace{-10pt}
\paragraph{SC Harmonization Framework:}
The encoder $\mathcal{F}_E$ encodes an input structural connectivity matrix $\mathbf{S}^c$ from site $c$ to a site-invariant embedding $f_E^c$, eliminating site-specific bias. Conversely, the site-classifier $\mathcal{F}_C$, a multi-layer perceptron (MLP), operates adversarially against $\mathcal{F}_E$ to predict the acquisition site $y^c$ of the input matrix. Thereby, $\mathcal{F}_C$ challenges $\mathcal{F}_E$ to produce an embedding that obscures the site information. This adversarial interplay is facilitated by a gradient reversal layer (GRL) \citep{ganin2015unsupervised}. During the forward pass, GRL acts as an identity function, and during backpropagation, it scales the inverted gradients by a dynamic factor $\lambda$, encouraging $\mathcal{F}_E$ to create site-agnostic embedding. 
In parallel, the site-mapper $\mathcal{F}_M$ maps the site-conditioned vector $v^c$, summarized in \textbf{Table \ref{tab:site_params}}, to a latent vector $f_M^c$ via an MLP. The latent-fusion module $\mathcal{F}_L$ combines the site-invariant embedding $f_E^c$ and the latent vector $f_M^c$, and the decoder $\mathcal{F}_D$ operates on the fused representation to output matrix $\hat{\mathbf{S}}^c$. 
During training, without traveling subjects, $v^c$ corresponds to the site of the input matrix and $\mathcal{F}_D$ aims to reconstruct the input matrix, \textit{i.e.} $\hat{\mathbf{S}}^c \simeq \mathbf{S}^c$. 
However, during validation and testing, connectivity matrices are harmonized to the highest b-value and spatial resolution by adjusting $v^c$ to the target site with the highest quality. $\mathcal{F}_D$ transforms the input matrix by blending subject-specific traits with target site-specific attributes. This generative modeling approach within our framework effectively harmonizes SCs across sites, minimizing variability while preserving key biological information for robust cross-site analyses.

The harmonization framework is trained by jointly optimizing the mean absolute error $\mathcal{L}_{\text{MAE}}$ between target and predicted connectivity matrices, and the categorical cross-entropy $\mathcal{L}_{\text{CE}}$ for the site-classifier, penalizing incorrect site classification. Since connectivity matrices are sparse, the MAE assigns higher weight to errors in predicting edge weights (\textit{i.e.} number of fibers) for existing connections compared to absent ones. 
Given a batch of $B$ input subjects, $\mathbf{S}_i$ and $\hat{\mathbf{S}}_i$ denoting the input and predicted connectivity matrices, and $y_i^c$ and $\hat{y}_i^c$ denoting the input and predicted site-labels for the $i^{th}$ subject, MAE and CE are computed as,
\[
\mathcal{L}_{\text{MAE}} = \frac{1}{B} \sum_{i=1}^{B} \left| \mathbf{S}_i - \hat{\mathbf{S}}_i \right|; \qquad
\mathcal{L}_{\text{CE}} = - \frac{1}{B} \sum_{i=1}^{B} \sum_{c=1}^{C} y_i^c \log\left( \frac{e^{\hat{y}_i^c}}{\sum_{k=1}^{C} e^{\hat{y}_i^k}} \right)
\]
where, $C$ is the number of site-classes. For CAE and GAE, an additional binary cross-entropy loss $\mathcal{L}_{\text{BCE}}$ is optimized enforcing the prediction of edge presence or absence. It serves as a regularizer alongside $\mathcal{L}_{\text{MAE}}$ and enhances the model's ability to capture brain connectivity structure.
\[
\mathcal{L}_{\text{BCE}} = - \frac{1}{B} \sum_{i=1}^{B} \left[ s_i^e \log(\sigma(\hat{s}_i^e)) + (1 - s_i^e) \log(1 - \sigma(\hat{s}_i^e)) \right]
\]
where, $s_i^e \in \{0,1\}$ is binary ground truth, $\hat{s}_i^e$ is predicted label, and $\sigma$ is the sigmoid activation.

\vspace{-10pt}
\paragraph{Latent Fusion Module:}
This section describes strategies for fusing site-invariant embedding $f_E$ from $\mathcal{F}_E$ and site-specific latent vector $f_M$ from $\mathcal{F}_M$ across different DL architectures.

For FAE, a concatenation operation along the embedding dimension directly fuses $f_E$ and $f_M$. 
For CAE and GAE, adaptive instance normalization (AdaIN) \citep{karras2019style} is used for fusion. Let $f_E \in \mathbb{R}^{B \times H \times W \times K}$ for 2D input and $f_E \in \mathbb{R}^{B \times N \times K}$ for graphs, where $B$ is batch size, $K$ is number of features, and $H, W$ are feature map dimensions for 2D, while $N$ is number of nodes for graphs. The mean and standard deviation of each feature $k$ are computed as,
\[
\mu^{bk} = \frac{1}{HW} \sum_{h=1}^{H} \sum_{w=1}^{W} f_E^{bhwk}; \quad
\sigma^{bk} = \Big[\frac{1}{HW} \sum_{h=1}^{H} \sum_{w=1}^{W} \left(f_E^{bhwk} - \mu_{bk}\right)^2\Big]^{1/2}; \qquad f_E \text{ as 2D input}
\]
\[
\mu^{bk} = \frac{1}{N} \sum_{n=1}^{N} f_E^{bnk}; \quad
\sigma^{bk} = \Big[\frac{1}{N} \sum_{n=1}^{N} \left(f_E^{bnk} - \mu_{bk}\right)^2\Big]^{1/2}; \qquad f_E \text{ as graph}
\]
Afterwards $f_E$ is feature-wise Z-score normalized for each subject using $\mu_{bk}$ and $\sigma_{bk}$. In parallel, the latent vector $f_M$ is processed by two fully connected networks to output parameter vectors $f_M^{s, bk}$ and $f_M^{t, bk}$ for feature-wise scaling and shifting of $f_E$, respectively. Formally the AdaIN operation is defined as,
\[
\text{AdaIN}(f_E^{bk}, f_M) = f_M^{s, bk} \cdot \frac{f_E^{bk} - \mu^{bk}}{\sigma^{bk}} + f_M^{t, bk}
\]
This formulation, based on AdaIN, ensures that the learned site-specific variations encoded in $f_M$ dynamically adjust the feature representations $f_E$ by aligning their statistical properties, effectively enabling adaptability while preserving the integrity of subject-specific information.

\paragraph{Fully Connected Autoencoder (FAE):}
FAE harmonizes the vectorized representation $\mathbf{s}^c \in \mathbb{R}^D$ of an input structural connectivity matrix $\mathbf{S}^c \in \mathbb{R}^{N \times N}$. The upper triangular elements of $\mathbf{S}^c$ are flattened into result in $\mathbf{s}^c$, following the strategy from the LR framework, where $D = (N^2 - N)/2$. $\mathbf{s}^c$ is fed into  $\mathcal{F}_E$, comprising fully connected layers, to produce a compressed embedding $f_E$. Next, $f_E$ is input into $\mathcal{F}_C$ for site-classification and concatenated with $f_M$ before passing into $\mathcal{F}_D$. Similar to $\mathcal{F}_E$, $\mathcal{F}_D$ includes fully connected layers, upsampling the fused embedding to reconstruct $\mathbf{s}^c$. The operation within the fully connected layers is expressed as, $\mathbf{s}^c_{l+1} = \psi\left(\text{Norm}(\textbf{W}_l \cdot \mathbf{s}^c_l + b_l)\right)$, where $\mathbf{s}^c$ is up/down-sampled from layer $l$ to $l+1$ using weight matrix $\textbf{W}_l$ and biases $b_l$. Here, $\psi$ denotes the non-linear activation and $\text{Norm}(\cdot)$ represents normalization layers, \textit{e.g.} BatchNorm or LayerNorm \citep{ioffe2015batch, ba2016layer}. Detailed architecture of FAE is provided in Supplementary Figure \ref{fig:FAE}.

\paragraph{Convolutional Autoencoder (CAE):}
CAE processes an input structural connectivity matrix $\mathbf{S}^c \in \mathbb{R}^{N \times N}$ as a single-channel image using a convolutional encoder-decoder architecture for harmonization. The encoder $\mathcal{F}_E$, based on ResNet \citep{he2016deep}, produces a compressed embedding $f_E \in \mathbb{R}^{H \times W \times K}$, where $K$ is the number of embedding channels, and $H$, $W$ are the spatial dimensions. 
The site-classifier $\mathcal{F}_C$ applies global average pooling across the spatial dimensions to convert $f_E$ into a vector $\in \mathbb{R}^{K}$ and utilizes fully connected layers to predict the site-class.
Concurrently, the site-mapper $\mathcal{F}_M$ produces $f_M \in \mathbb{R}^{K}$, which serves as input to the latent fusion module $\mathcal{F}_L$, incorporating AdaIN. 
The decoder $\mathcal{F}_D$ comprises sequential deconvolution layers, each including a transposed convolutional layer, convolutional layers, and a $\mathcal{F}_L$ module, to reconstruct $\mathbf{S}^c$ from $f_E$. Notably, $\mathcal{F}_L$ modules integrate site-conditioning deeply within each deconvolution layer, enabling the generation of site-specific connectivity matrices. The final layer of $\mathcal{F}_D$ uses a $1 \times 1$ convolution layer to output the reconstructed $\mathbf{S}^c$. Detailed architecture of FAE is provided in Supplementary Figure \ref{fig:CAE}.

\vspace{-20pt}
\paragraph{Graph Autoencoder (GAE):}
GAE harmonizes the graph representation of an input structural connectivity matrix $\mathbf{S}^c \in \mathbb{R}^{N \times N}$. The node feature matrix $\mathbf{X} \in \mathbb{R}^{N \times N}$ and the adjacency matrix $\mathbf{A} \in \mathbb{R}^{N \times N}$ of the graph are identity matrix and $\mathbf{S}^c$, respectively. 
The encoder $\mathcal{F}_E$ consists of $M_N$ Chebyshev graph convolution layers (ChebConv) \citep{kipf2016semi} and produces a compressed embedding $f_E \in \mathbb{R}^{N \times K}$, where $K$ is the embedding dimension.

A ChebConv layer begins with computing the normalized graph Laplacian $\mathbf{L}$ as follows where, $\mathbf{I}_N$ is a $N \times N$ identity matrix and $\mathbf{D}$ is the diagonal degree matrix with elements $\mathbf{D}_{uu} = \sum_{v=1}^N \mathbf{A}_{uv}$. 
\[
\mathbf{L} = \mathbf{I}_N - \mathbf{D}^{-\frac{1}{2}} \mathbf{A} \mathbf{D}^{-\frac{1}{2}}
\]
The Laplacian $\mathbf{L}$ can be decomposed as $\mathbf{L} = \mathbf{U} \mathbf{\Lambda} \mathbf{U}^\top$, where $\mathbf{U} \in \mathbb{R}^{N \times N}$ is the matrix of eigenvectors and $\mathbf{\Lambda} = \text{diag}([\lambda_1, \lambda_2, \dots, \lambda_N])$ is a diagonal matrix containing the eigenvalues.
A signal $\mathbf{x} \in \mathbb{R}^N$ defined on the graph's vertices can be transformed into the spectral domain as $\hat{\mathbf{x}} = \mathbf{U}^\top \mathbf{x}$. A convolution operation on $\mathbf{x}$ with a graph filter $g$ in the vertex domain is equivalent to multiplication in the spectral domain as, 
\[
\mathbf{x} * g = \mathbf{U} \hat{g}(\mathbf{\Lambda}) \mathbf{U}^\top \mathbf{x} = \hat{g}(\mathbf{L}) \mathbf{x}
\]
Here, $\hat{g}(\mathbf{\Lambda})$ denotes the filter function applied to the eigenvalues of $\mathbf{L}$, and $\hat{g}(\mathbf{L})$ is its matrix polynomial form. To ensure that $\hat{g}(\mathbf{L})$ is $M$-localized (\textit{i.e.}, dependent only on nodes up to $M$ hops away) and to avoid the computational cost of eigen-decomposition, it is approximated using Chebyshev polynomials of order $M$, given as,
\[
g_\theta(\mathbf{L}) = \sum_{m=0}^{M} \theta_m T_m(\tilde{\mathbf{L}})
\]
where, 
$T_m(x)$ are Chebyshev polynomials of degree $m$,
$\tilde{\mathbf{L}} = \frac{2}{\lambda_{\max}} \mathbf{L} - \mathbf{I}_N$ is the rescaled graph Laplacian,
$\lambda_{\max}$ is the largest eigenvalue of $\mathbf{L}$, and
$\theta_m$ are the trainable parameters.
The Chebyshev polynomials \( T_m(x) \) are defined recursively as:
\[
T_0(x) = 1, \quad T_1(x) = x, \quad T_m(x) = 2x T_{m-1}(x) - T_{m-2}(x), \quad m \geq 2
\]
Using the above formulation and approximation, a ChebConv layer updates the node features $\mathbf{X}^{(m)} \in \mathbb{R}^{N \times d^{(m)}}$ from layer $m$ to layer $m+1$ as, 
\[
\mathbf{X}^{(m+1)} = \sigma \left( \text{Norm} \left( \sum_{m=0}^{M} \theta_m T_m(\tilde{\mathbf{L}}) \mathbf{X}^{(m)} \right) \right)
\]
where, $d^{(m)}$ is the feature dimension at layer $m$,
$\sigma$ is a non-linear activation such as ReLU, and
\text{Norm} denotes a normalization operation (\textit{e.g.}, batch normalization). 

Following $M_N$ ChebConv layers, we get $f_E \in \mathbb{R}^{N \times K}$. The site-classifier $\mathcal{F}_C$ aggregates $f_E$ across nodes using global average- and max-pooling, and subsequently concatenates the aggregated features and passed through a feedforward network to predict site-class. 
Concurrently, the site-mapper $\mathcal{F}_M$ produces $f_M \in \mathbb{R}^K$.
The decoder $\mathcal{F}_D$ sequentially includes multiple fully connected layers and multiple ChebConv layers to reconstruct $\mathbf{S}^c$. All fully connected and ChebConv layers include an upsampling operation and a latent fusion module $\mathcal{F}_L$, incorporating AdaIN, analogous to CAE. Detailed architecture of FAE is provided in Supplementary Figure \ref{fig:GAE}.

As in the case of LR, some post-processing steps were applied to the predicted connectomes. For the CAE and GAE, since the outputs were matrices, each matrix was first symmetrized and its diagonal entries were set to zero. Subsequently, for all DL models, any negative estimated values were clipped to zero, and the resulting values were rounded to the nearest integer to reflect the inherently discrete nature of fiber counts.

\subsection{Evaluation metrics}

The SC harmonization methods are evaluated on the test set across various metrics. They compare the quality of the harmonized structural connectivity matrices from the lowest-quality SCs (b-value 1000 and resolution 2.3), denoted as $\mathcal{S}^{\text{low}}_{\text{h}}$, against the structural connectivity matrices from the highest-quality SCs (b-value 3000 and resolution 1.25), denoted as $\mathcal{S}^{\text{high}}$, across all test subjects. These metrics can be broadly categorized as, 
\textbf{(1)} Edge-level accuracy, 
\textbf{(2)} Topological preservation, 
\textbf{(3)} Individuality retention,
\textbf{(4)} Demographic prediction, and
\textbf{(5)} Domain-invariant embedding space.
The lower-bounds of these metrics are computed by comparing the lowest-quality SCs without any harmonization $\mathcal{S}^{\text{low}}_{\text{nh}}$ against $\mathcal{S}^{\text{high}}$. 
The upper-bounds of these metrics are computed by comparing $\mathcal{S^{\text{high}}}$ across the test set and an independent retest set for the same subjects. The upper-bounds represent the optimal achievable scores while considering the inherent variability in the data. To note, 42 subjects constitute the retest set.

\textbf{\textit{Edge-level accuracy}} evaluates how accurately the edge weights of $\mathcal{S}^{\text{low}}_{\text{h}}$ are preserved compared to $\mathcal{S}^{\text{high}}$. Preservation is quantified using the \textbf{mean absolute error (MAE)} and the \textbf{Pearson correlation (PC)}. Further \textbf{binary MAE (BMAE)} is calculated on the binary edges to measure the preservation of structural connectivity.

\textbf{\textit{Topological preservation}} evaluates how accurately the nodal scores are preserved between $\mathcal{S}^{\text{low}}_{\text{h}}$ and $\mathcal{S}^{\text{high}}$. The nodal scores measure \textbf{nodal strength (NS)}, \textbf{closeness centrality (CC)}, \textbf{clustering coefficient (CLC)}, \textbf{local efficiency (LE)} and \textbf{eigenvalue (EV)} for all the nodes in a connectivity matrix. The metrics are calculated as the mean absolute error between the nodal scores from $\mathcal{S}^{\text{low}}_{\text{h}}$ and $\mathcal{S}^{\text{high}}$. For EV, mean absolute error is computed between the sorted eigenvalues.

\textbf{\textit{Individuality retention}} evaluates how accurately individual variability is  preserved between $\mathcal{S}^{\text{low}}_{\text{h}}$ and $\mathcal{S}^{\text{high}}$. Preservation is quantified using \textbf{fingerprinting accuracy (FA)} and \textbf{identifiability difference (ID)} \citep{amico2018quest}.
To compute FA, first pair-wise distances are calculated between the connectivity matrices $\mathbf{S}^{\text{low}}_{\text{h},i} \in \mathcal{S}^{\text{low}}_{\text{h}}$ and $\mathbf{S}^{\text{high}}_j \in \mathcal{S}^{\text{high}}$, resulting in $\mathcal{P} \in \mathbb{R}^{|\mathcal{S}| \times |\mathcal{S}|}$ matrix. The distance is measured as the mean absolute error between the upper triangular vectors from $\mathbf{S}^{\text{low}}_{\text{h},i}$ and $\mathbf{S}^{\text{high}}_j$. Next, FA is calculated as the accuracy of obtaining the minimum distance on the diagonal of $\mathcal{P}$.
ID is measured as the difference between the mean inter-subject distance and the mean intra-subject distance from $\mathcal{P}$. A higher ID is better as it signifies that the SCs of the same individual are closer following the harmonization, compared to the SCs of others.

\textbf{\textit{Demographic prediction}} measures the accuracies of \textbf{gender prediction (GP)} and \textbf{age prediction (AP)}. 
First, gender and age predictive models are trained using the training set of 848 structural connectivity matrices from the highest-quality SCs. Specifically, a RBF-kernel SVM and a linear regression model for gender and age prediction, respectively. Next, the models are evaluated on $\mathcal{S}^{\text{low}}_{\text{h}}$, $\mathcal{S}^{\text{low}}_{\text{nh}}$, and $\mathcal{S}^{\text{high}}$, in terms of accuracy and Pearson correlation, to compute the metrics for the harmonization method, the lower-bound, and the upper-bound. 
 

\textbf{\textit{Spyder plot:}}
For comparing the deep learning models against each metric, a spyder plot was plotted with metrics as MAE, BMAE, EV, NS, CC, CLC, LE, FA, ID, AP and GP. All these features were first normalized to have values between 0 and 1 and then some of them were transformed such that the higher the better. The farther the values from the center, the better the model for that particular metric.

\textbf{\textit{Qualitative metrics:}}
We qualitatively assess the encoder's embeddings using t-SNE \citep{van2008visualizing}, color-coded by site and subject. For a site-invariant encoder, embeddings across sites should ideally be mixed. Additionally, preserving subject-specific traits implies embeddings for each subject across sites should cluster together. Achieving both site mix-up and subject clustering reflects an ideal embedding space.


\vspace{-20pt}
\subsection{Implementation details}

In this subsection we present the implementation details for data handling and the experimented hyperparameters for modeling FAE, CAE, and GAE.

The connectivity matrices undergo log-normalization prior to input into the FAE, defined as, $\mathbf{S}_i^c = \log_e(\mathbf{S}_i^c + 1)$. In contrast, CAE and GAE use the raw connectivity matrices.

The latent fusion module for CAE and GAE include AdaIN with GRL scaling parameter $\lambda$. For training stability, $\lambda$ is progressively increased from 0 to 1 over the first 100 epochs using the scheduler: $\lambda = \frac{2}{1 + \exp(-\gamma \cdot p)} - 1$, where $p$ is the normalized training progress. Beyond the 100$^{th}$ epoch, $\lambda$ remains close to 1. Following recommendations by \citep{ganin2016domain}, $\gamma$ is set to 10.

FAE, CAE, and GAE are trained with batch size 32. FAE and GAE are optimized using Adam optimizer \citep{kingma2014adam} with initial learning rate $1e^{-2}$ for encoder and decoder, and $1e^{-3}$ for site-mapper and site-classifier. The CAE encoder is a ResNet18 architecture. CAE is optimized using Adam with initial learning rate $1e^{-3}$ for encoder, decoder, site-mapper and site-classifier.
For all models, the learning rate was reduced by a factor of 0.9 if the training loss did not improve for 5 consecutive epochs, with an improvement threshold of $1e^{-5}$.
%
In $\mathcal{L}_{\text{MAE}}$, errors in predicting edge weights for existing connections is weighted $2.5\times$ compared to non-connections.
All models are trained for 1000 epochs.

To select the best model, MAE and FA metrics are evaluated during validation by comparing the predicted SCs for the lowest quality input to the highest-quality target SCs. The epoch yielding the optimal balance of lower MAE and higher FA is selected as the final model.

\begin{figure}[t]
\centerline{\includegraphics[width=0.5\textwidth]{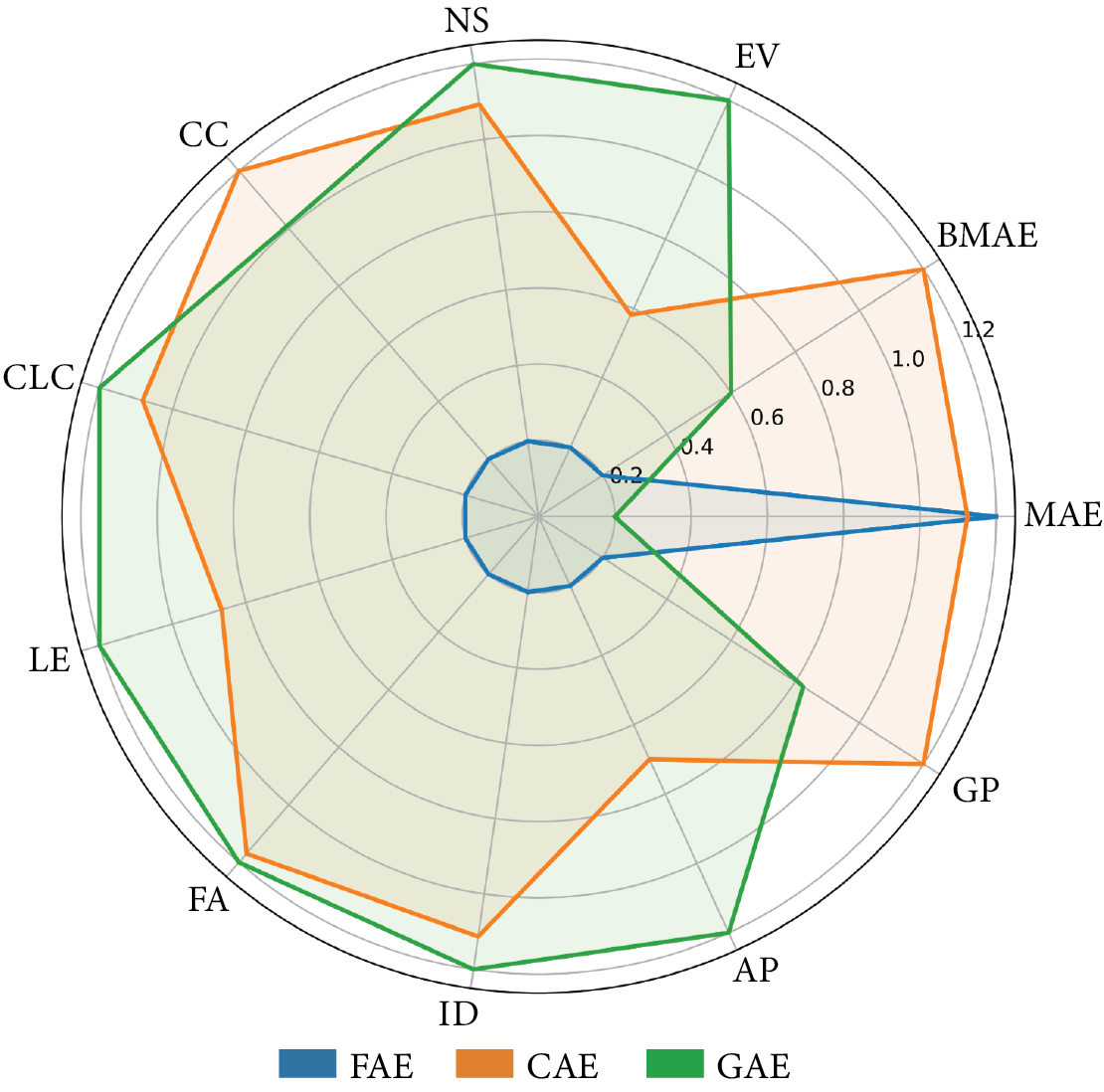}}
\caption{\textbf{Spyder plot comparing deep SC harmonization methods.} Metrics are normalized to a [0, 1] range, with MAE, BMAE, EV, NS, CC, CLC, and LE inverted for consistency, where higher values indicate better performance.}
\label{fig:spyder}
\vspace{-1.0em}
\end{figure}

\begin{table}[!t]
    \centering
    \setlength{\tabcolsep}{4pt}
    \renewcommand{\arraystretch}{1.0}
    \begin{tabular}{c|c|c|c|c|c|c}
        \toprule
        \textbf{} & & \multicolumn{4}{c|}{\textbf{SC harmonization methods}} &  \\ 
        \cline{3-6}
        \diagbox[width=7em]{\textbf{Metrics}}{\textbf{Methods}} & \textbf{Lower-bound} & \textbf{LR} & \textbf{FAE} & \textbf{CAE} & \textbf{GAE} & \textbf{Upper-bound} \\
        \midrule
        \multirow{2}{*}{MAE ($\downarrow$)} & $8.18$ & $4.92$ & $5.83$ & $5.89$ & $6.63$ & $4.08$ \\
         & $\pm 0.99$ & $\pm 0.38$ & $\pm 0.58$ & $\pm 0.54$ & $\pm 0.49$ & $\pm 1.38$ \\
        \midrule
        \multirow{2}{*}{BMAE ($\downarrow$)} & $0.19$ & $0.16$ & $0.20$ & $0.15$ & $0.18$ & $0.14$ \\
         & $\pm 0.01$ & $\pm 0.01$ & $\pm 0.01$ & $\pm 0.01$ & $\pm 0.01$ & $\pm 0.02$ \\
        \midrule
        \multirow{2}{*}{PC ($\uparrow$)} & $0.94$ & $0.98$ & $0.95$ & $0.95$ & $0.95$ & $0.98$ \\
         & $\pm 0.01$ & $\pm 0.003$ & $\pm 0.01$ & $\pm 0.01$ & $\pm 0.01$ & $\pm 0.01$ \\
        \midrule

        \multirow{2}{*}{NS ($\downarrow$)} & $1869.97$ & $428.93$ & $737.78$ & $652.29$ & $641.99$ & $394.12$ \\
         & $\pm 303.77$ & $\pm 129.34$ & $\pm 235.97$ & $\pm 200.30$ & $\pm 177.98$ & $\pm 189.47$ \\
        \midrule
        \multirow{2}{*}{CC ($\downarrow$)} & $30.78$ & $6.23$ & $13.52$ & $7.79$ & $8.89$ & $5.93$ \\
         & $\pm 5.83$ & $\pm 2.54$ & $\pm 6.54$ & $\pm 2.67$ & $\pm 3.36$ & $\pm 3.19$ \\
        \midrule
        \multirow{2}{*}{CLC ($\downarrow$)} & $5.70$ & $3.09$ & $5.84$ & $4.05$ & $3.81$  & $1.91$ \\
         & $\pm 0.69$ & $\pm 0.54$ & $\pm 0.82$ & $\pm 0.71$ & $\pm 0.36$  & $\pm 0.43$ \\
        \midrule
        \multirow{2}{*}{LE ($\downarrow$)} & $9.85$ & $6.09$ & $12.47$ & $8.4$ & $6.36$ & $3.32$ \\
         & $\pm 1.18$ & $\pm 1.12$ & $\pm 1.74$ & $\pm 1.49$ & $\pm 0.56$ & $\pm 0.74$ \\
        \midrule
        \multirow{2}{*}{EV ($\downarrow$)} & $1869.39$ & $288.10$ & $522.61$ & $518.14$ & $462.46$ & $248.31$ \\
         & $\pm 304.34$ & $\pm 177.58$ & $\pm 322.27$ & $\pm 281.43$ & $\pm 248.72$ & $\pm 184.40$ \\
        \midrule
        
        FA ($\uparrow$) & 0.20 & 1.0 & 0.01 & 0.96 & 0.99 & 1.0 \\
        \midrule
        ID ($\uparrow$)& 1.55 & 2.37 & 0.013 & 2.31 & 2.53  & 3.89 \\
        \midrule
        
        GP ($\uparrow$) & 0.80 & 0.85 & 0.75 & 0.83 & 0.80  & 0.87 \\
        \midrule
        AP ($\uparrow$) & 0.18 & 0.18 & 0.15 & 0.16 & 0.17 & 0.29 \\
        \bottomrule
    \end{tabular}
    \caption{\textbf{Quantitative comparison of SC harmonization methods, LR, FAE, CAE, and GAE, for harmonizing the lowest-quality SCs [bvalue=1000; resolution=2.3] to the highest quality SCs [bvalue=3000; resolution=1.25].} Lower-bound is defined by comparing the lowest-quality SCs to the highest quality SCs without any harmonization. Upper-bound is defined by comparing the highest quality SCs across 42 test-and-retest subjects except for AP and GP. For these metrics, the upper bounds are calculated using the highest quality test SCs.}
    \label{tab:metrics}
\end{table}

\begin{figure}[!t]
\centerline{\includegraphics[width=0.95\textwidth]{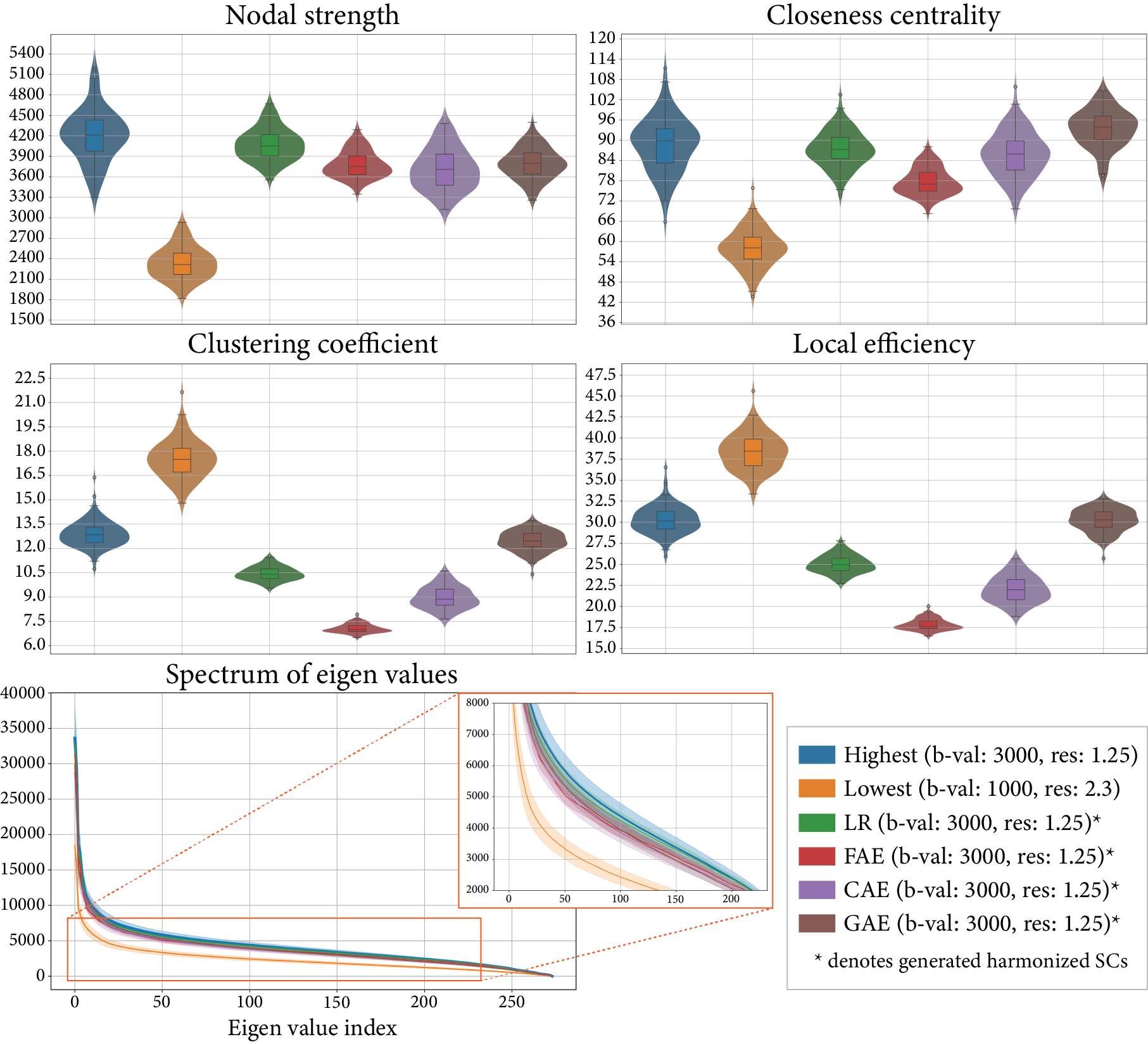}}
\caption{Distribution of topological preservation metrics across test subjects for the highest-quality SCs, the lowest-quality SCs, and the lowest- to the highest-quality harmonized SCs generated by LR, FAE, CAE, and GAE.}
\label{fig:graphmetrics}
\vspace{-2em}
\end{figure}

\section{Results}
This section provides both quantitative and qualitative evaluations of methods across multiple criteria. Figures \ref{fig:spyder} and \ref{fig:harmonization} summarize the quantitative and qualitative results, respectively.

\subsection{Quantitative evaluation}
A comparative summary of FAE, CAE, and GAE is illustrated in Figure \ref{fig:spyder}. FAE performs the worst overall, except in MAE. CAE excels in edge-level metrics (MAE, BMAE), while GAE consistently outperforms CAE in higher-level graph metrics (excluding CC) and identifiability. For demographic prediction, CAE and GAE perform similarly, with CAE leading in GP and GAE slightly ahead in AP. Specific metric details are discussed below.

\subsubsection{Edge-level accuracy}
As summarized in Table \ref{tab:metrics}, LR achieves the lowest MAE, while GAE reports the highest, and FAE and CAE perform comparable. Despite these variations, all models substantially reduce MAE relative to the unharmonized lower-bound, demonstrating their effectiveness in mitigating acquisition-related distortions. Moreover, the model-induced errors are close to the upper-bound defined by test-retest reliability across subjects, underscoring their potential in harmonization tasks.
However, BMAE reveals a different pattern. CAE achieves the best result, while FAE performs the worst, exceeding even the unharmonized lower-bound. This suggests that while FAE can learn edge weight distributions, it struggles to preserve structural topology. Conversely, CAE excels in capturing localized spatial patterns and retaining structural integrity. LR and GAE fall between these extremes. The test-retest BMAE sets an upper-bound, indicating all models—except FAE—perform reasonably well in preserving connectomic structure.
For Pearson correlation, LR achieves the highest performance, closely matching the test-retest upper-bound. Deep learning models yield slightly lower yet consistent results, still surpassing the unharmonized lower-bound, indicating their capability for capturing global inter-regional relationships.
In summary, LR outperforms DL models on MAE and Pearson correlation, reflecting strength in capturing global edge-weight patterns, while CAE excels at preserving structural properties via BMAE. These differences emphasize distinct model capacities in harmonizing connectivity data, which are further elaborated in the Discussion section.

\subsubsection{Topological preservation}
As shown in Table \ref{tab:metrics}, the mean absolute error in EV, NS, CC, CLC, and LE are the lowest for LR, and the highest for FAE. Except for CC, GAE consistently outperforms CAE across all other metrics.
A notable observation is that for EV, NS, and CC, LR performs very close to upper-bound. However, for CLC and LE, the differences between LR and the upper-bound are relatively large, and GAE achieves comparable results to LR.

Figure \ref{fig:graphmetrics} presents violin plots illustrating the distribution of nodal values across test subjects, unlike the metrics in Table \ref{tab:metrics}, which present an aggregated view of the nodal differences between the highest-quality SCs and the harmonized SCs across all subjects.
For the violin plots, we compute these node-level metrics for each subject. This produces a matrix of shape $\mathbb{R}^{M \times N_s \times N}$, where $M$, $N_s$, and $N$ are the number of metrics, number of test subjects, and number of nodes. Then we average across the nodes to compute an aggregated matrix $\mathbb{R}^{M \times N_s}$, which signifies the distribution of metrics across the subjects. The violin plots present this aggregated matrix for the highest-quality SCs ($\mathcal{S}^{\text{high}}$), the lowest-quality SCs ($\mathcal{S}^{\text{low}}$), and the harmonized SCs generated by LR, FAE, CAE, and GAE. For each metric, distribution similarity to $\mathcal{S}^{\text{high}}$ indicates topological preservation. 
For NS, LR closely matches $\mathcal{S}^{\text{high}}$, while FAE, CAE, and GAE perform similarly. For CC, both LR and CAE align well with $\mathcal{S}^{\text{high}}$, while GAE slightly overestimates. Across both metrics, all models significantly outperform $\mathcal{S}^{\text{low}}$.
The notable results appear for CLC and LE, where GAE almost matches $\mathcal{S}^{\text{high}}$. In contrast, other models overcorrect, yielding lower mean than $\mathcal{S}^{\text{high}}$, with FAE performing the worst. 
This contradicts Table \ref{tab:metrics}, where LR outperforms GAE in terms of node-level CLC and LE. The observation suggests that GAE is better at maintaining subject-level connectivity than LR, whereas LR is better at maintaining node-level accuracy.

For the eigen values plot in Figure \ref{fig:graphmetrics}, we compute the eigen values for each subject, producing a $\mathbb{R}^{N_s \times N}$ matrix. We aggregate across the subjects and plot the means and standard deviations for each eigen value. All harmonization methods nicely match the highest-quality SCs, with LR matching the most. Further, all methods significantly exceed the trend of the lowest-quality SCs.

\subsubsection{Individuality retention}
As shown in Table \ref{tab:metrics}, FA and ID values improve significantly after harmonization for all methods except FAE, which shows a notable decline in performance. For FA, LR achieves the perfect accuracy (100\%), closely followed by GAE at 99\%. Regarding ID, GAE performs the best, followed by LR and CAE. However, none of the methods reach ID values comparable to the optimal upper-bound metrics.

\subsubsection{Demographic prediction}
For GP, LR achieves the highest accuracy (Table \ref{tab:metrics}), only 2\% below the upper-bound, while CAE improves post-harmonization and GAE remains stable. In contrast, FAE shows a decline in performance after harmonization. For AP, all models except LR experience mild performance declines post-harmonization, and GAE performs comparable to LR. Notably, none of the models approach upper-bound performance for this metric. Interestingly, despite FAE's poor performance on FA and ID, it performs relatively better on demographic prediction tasks like GP and AP. This suggests FAE may capture features more relevant to demographic attributes than to individual-specific identification.


\begin{sidewaysfigure}
\setlength{\abovecaptionskip}{-1pt}  
\centerline{\includegraphics[width=\textwidth]{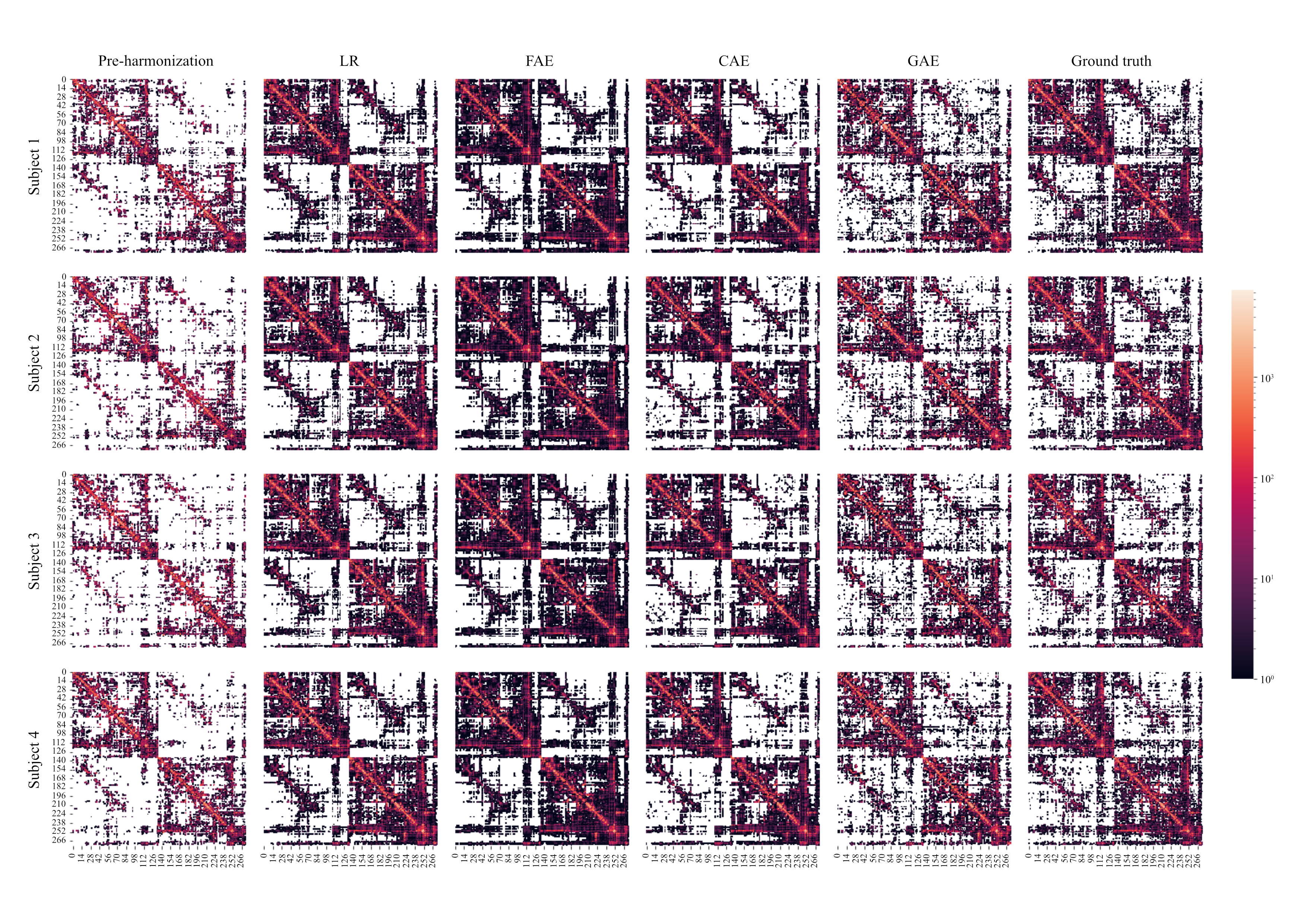}}
\caption{
\textbf{Qualitative comparison of structural connectivity matrices.} 
The lowest-quality (Pre-harmonization) and the highest-quality (Ground truth) connectivity matrices are shown for four subjects, along with the predicted connectivity matrices from four models—Linear Regression (LR), Fully connected Autoencoder (FAE), Convolutional Autoencoder (CAE), and Graph Autoencoder (GAE)—for harmonizing lowest- to highest-quality. The colorbar signifies the number of connections and white indicates the absence of connections.}
\label{fig:harmonization}
\end{sidewaysfigure}

\subsection{Qualitative evaluation}
As illustrated in Figure \ref{fig:harmonization}, the predicted structural connectivity matrices generated by all models—LR, FAE, CAE, and GAE—visually resemble the ground truth more closely than the unharmonized input matrices, indicating their ability to mitigate acquisition-related discrepancies. However, an intriguing pattern emerges: all models except GAE tend to overcorrect acquisition effects, notably in FAE. This suggests that while harmonization reduces site-related effects, some models inadvertently amplify protocol-specific distortions.

Further scrutiny reveals differences in preserving subject-level individuality. For example, when comparing rows for \textit{Subject 1}, \textit{Subject 2}, \textit{Subject 3} and \textit{Subject 4}, the harmonized outputs of all models except GAE display increased homogeneity, emphasizing dominant connections while overlooking subtle ones—a shortcoming most evident in FAE. In contrast, GAE shows stronger preservation of individual patterns, reconstructing less prominent yet probably biologically relevant connections. While GAE’s outputs may appear lighter overall, they contain richer detail in low-weight connections, potentially capturing subtle inter-regional relationships that contribute to individual variability. This ability to preserve fine-grained features highlights GAE’s advantage in retaining biological signals during harmonization.

\begin{figure}[!t]
\centerline{\includegraphics[width=0.9\textwidth]{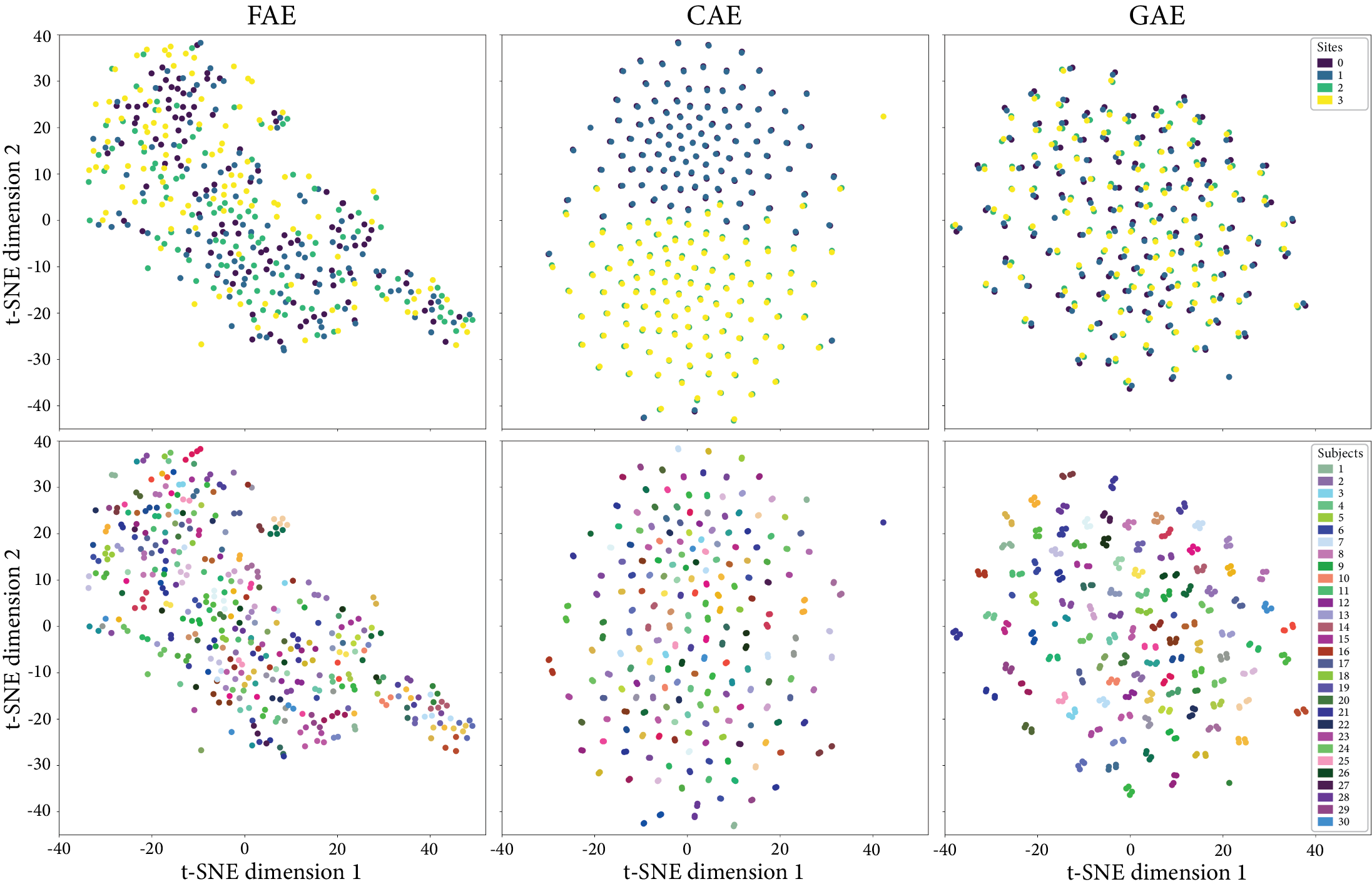}}
\caption{t-SNE visualization of encoder embeddings from the deep SC harmonization framework for FAE, CAE, and GAE. The embeddings are color-coded by site and subject.}
\label{fig:t-SNE}
\end{figure}

\textbf{\textit{t-SNE visualization:}}
Figure \ref{fig:t-SNE} highlights key differences among the methods. FAE excels in site-level data mixing but struggles with subject-level clustering. CAE poorly mixes site data, forming two distinct clusters—Sites 0 and 1 grouping together, and Sites 2 and 3 clustering another. Its subject-level clustering is also limited, observed in only a few cases. GAE achieves the best balance, effectively mixing site data while maintaining strong subject-level clustering.



\section{Discussion}

This study presents an innovative deep learning (DL) framework featuring three architectures—fully connected autoencoder (FAE), convolutional autoencoder (CAE), and graph convolutional autoencoder (GAE)—designed for structural connectome (SC) harmonization across diverse acquisition parameters and sites without relying on traveling subjects. While the FAE builds upon prior work \citep{newlin2024learning}, we introduce novel CAE and GAE models specifically tailored to this task. Unlike the top-performing linear regression (LR) approach \citep{patel2024modeling}, the proposed DL methods employ a site-conditioned adversarial network, eliminating the need for explicit acquisition metadata. Our comprehensive evaluation highlights the methods' ability to reduce site bias while preserving critical biological information, emphasizing the importance of model architecture in maintaining SC properties essential for downstream analyses.

Methods were evaluated across edge-level accuracy, topological preservation, individuality, demographic prediction, and domain-invariant learning in Table \ref{tab:metrics}. LR consistently achieved superior performance across most metrics, except for binary matrix mean absolute error (BMAE). Its advantage likely stems from access to detailed acquisition parameters during training, enabling precise modeling of generative factors underlying SC construction. However, in practical multi-site scenarios where such acquisition metadata is often unavailable or too complex to model, our DL methods offer a compelling alternative by achieving harmonization purely from site identity labels without explicit parameter knowledge. 

Noticeable performance variations across metrics were observed across DL architectures. FAE excelled in predicting precise edge weights but performed worst on BMAE. In contrast, CAE achieved the lowest BMAE, followed by GAE. These differences likely arise from architectural differences and training objectives. FAEs, unconstrained by graph structure and possessing a higher parameter count, may better capture edge weight distributions but struggles with modeling connection existence. In contrast, CAE's and GAE's convolutional operations, favoring neighborhood feature averaging, enhance detection of connection existence but at the cost of edge weight precision. The averaging effect is more pronounced in GAE, that aggregates information across the graph's structure rather than relying only on local spatial proximity. However, the averaging explains CAE's and GAE's superior BMAE, as it imposes a constraint detecting connections between nodes with shared neighbors over connections between unrelated nodes. CAE even slightly outperformed LR in BMAE, followed by GAE, suggesting that local spatial context offers a marginal advantage in detecting edge existence beyond acquisition parameters alone. GAE's slightly lower BMAE than CAE, despite both employing convolutions, reflects its stronger neighborhood bias, occasionally leading to spurious or missed connections. Overall, neighborhood context appears crucial for predicting edge existence but less ideal for precise weight prediction. Regarding Pearson correlation, which assesses linear relationships rather than exact magnitudes, all models performed similarly, indicating they capture the overall trend of weights to a comparable degree.

For graph metric preservation, GAE clearly outperformed CAE and FAE. Flattening the connectome as a vector, FAE failed to capture network topology, underperforming across all graph metrics. CAE surpassed GAE only for closeness centrality. GAE, specifically tailored for graph-structured data, learns how well a node is connected within the network (reflected by nodal strength) and captures the overall network organization (reflected by eigenvalues). Additionally, GAE's localized processing enables the detection of local triangle density (clustering coefficient) and the efficiency of local information exchange (local efficiency). GAE's slight overestimation of closeness centrality likely stems from predicted spurious connections, which reduces shortest path lengths between nodes and induces an increased closeness centrality. CAE’s local spatial filters captured some graph properties but struggled with connections between spatially distant but topologically related nodes. This highlights the advantage of true graph-based architectures for modeling complex, non-local connectome organization.

Regarding individuality retention (Table \ref{tab:metrics}, Figure \ref{fig:t-SNE}), GAE outperformed both CAE and FAE. GAE’s latent space disentangled site from subject information (Figure \ref{fig:t-SNE}), effectively clustering subjects regardless of the acquisition site. Furthermore, the harmonized matrices from GAE across subjects (Figure \ref{fig:harmonization}) exhibit distinct qualitative differences, indicating GAE's ability to capture subtle  subject-specific patterns. Such disentanglement likely underpins GAE's superiority in maintaining low intra-subject and high inter-subject distances, contributing to high fingerprinting accuracy (FA) and identifiability (ID). Though CAE retained some site bias in latent space (clustering sites into two groups), it managed to capture highly discriminative subject-specific features, as reflected in its strong FA and ID. Meanwhile, FAE, despite effectively mixing sites, failed to form clear subject clusters, resulting in inferior FA and ID performance.

In demographic prediction, CAE slightly improved gender classification post-harmonization (80\% to 83\%), GAE maintained performance (80\%), and FAE declined (80\% to 75\%). This suggests CAE's local spatial learning may better preserve or enhance gender-relevant features. GAE’s stable performance indicates its prioritization of individual over gender-specific signals. For age prediction, all models showed marginal decreases in Pearson correlation post-harmonization, suggesting reasonable preservation of age-related connectivity patterns. Notably, the connectivity differences across the age-groups appears to be subtle, as the upper-bound including high-quality SCs also struggles for age prediction. This is likely related to the narrow age range of the dataset (22-35 years old). Interestingly, despite weak fingerprinting, FAE performed relatively well in demographic tasks. This suggests that age and gender may correlate more strongly with global statistical properties like edge weight distribution, whereas fingerprinting requires fine-grained local and global details. Overall, these results indicate that different connectivity patterns are associated with distinct demographic variables, and each model possesses a unique capacity to capture these variations.

Overall, although DL methods trailed LR in absolute performance, they offer the critical advantage of harmonizing without relying on acquisition metadata. The choice of optimal DL model for harmonization should align with downstream goals: FAE may suffice for edge-weight regression tasks; CAE is better suited for demographic prediction and edge existence detection; and GAE is optimal for tasks requiring preservation of fine-grained individual and topological features. Graph-based models like GAE are especially crucial for applications which model complex topological organization of brain networks, such as individualized neuroscience and clinical research.

\subsection{Limitations and Future Work}
As discussed, while the DL methods—particularly CAE and GAE—show competitive performance compared to LR and  the upper-bound, they are not without limitations. CAE’s reliance on local spatial proximity overlooks critical long-range connections, a key drawback when modeling inherently graph-structured data like SCs. Consequently, future work should prioritize advancing the GAE methods. Specifically, improving GAE's edge prediction accuracy and demographic prediction performance without compromising fingerprinting and topological preservation. Enhancing GAE with more sophisticated GNN architectures that incorporate edge-edge convolutions \citep{zhang2019graph}, in addition to node-node interactions, could better capture the adjacency structure. Further, adopting complex frameworks such as Star Graph-GANs \citep{guo2022deep, wang2018graphgan, fan2019labeled, choi2018stargan}
, which use discriminators alongside graph generators, may impose stronger constraints during graph generation. Furthermore, incorporating loss functions that penalize spurious edges and reward accurate long-range connections warrants exploration. Beyond graph generation, analyzing and interpreting the GAE’s latent space for specific downstream tasks like demographic prediction and fingerprinting could offer valuable insights. Finally, extending the architecture to support variable node counts would enable super-resolution graph generation \citep{isallari2021brain}, allowing the creation of new nodes and connections, and broadening its applicability across a wider range of neuroimaging tasks. 
Also, this study's simulated design allowed for the controlled manipulation of two acquisition parameters and provided ground truth for validating the results. However, the generalizability of these findings—particularly in the context of applying GNNs to structural connectomes derived from multi-site data with varying acquisition parameters—remains to be fully established. Future research should focus on evaluating the framework's efficacy in mitigating scanner-related biases while reliably identifying group- or subject-level differences in more complex, real-world settings.


\section{Conclusion}
This work proposed a deep learning framework for structural connectome harmonization, evaluating three deep architectures—fully connected autoencoder (FAE), convolutional autoencoder (CAE), and graph convolutional autoencoder (GAE)—against a top-performing linear regression (LR) baseline. While LR achieved higher performance across most metrics by explicitly modeling acquisition parameters, its reliance on detailed acquisition metadata limits its practicality in real-world multi-site studies. In contrast, the proposed deep learning methods, particularly the GAE, harmonized connectomes purely from site identity, learning site-specific differences directly from the data without requiring acquisition details. The framework incorporated an adversarial site-classifier to enforce site-invariant encoding and a site-mapper to condition decoding on site-specific information. Results emphasized the importance of aligning model design with the data's inherent structure and harmonization goals. Specifically, the GAE—designed for graph-structured data—proved most effective at preserving both fine-grained individual characteristics and global topological properties critical for downstream tasks such as fingerprinting and network analysis. Furthermore, the GAE successfully disentangled site effects from subject identity in the latent space, reinforcing its suitability for structural connectome generation. Overall, these findings suggest that graph-aware deep learning approaches are particularly well-suited for harmonizing structural connectomes in applications demanding robust domain generalization and subject-level integrity, including personalized neuroscience and clinical research.

\begin{table}[!h]
    \centering
    \renewcommand{\arraystretch}{1.0}
    \begin{tabular}{c|c|c|c}
        \toprule
        \textbf{Abbreviation} & \textbf{Full Form} & \textbf{Abbreviation} & \textbf{Full Form} \\ 
        \midrule
        FAE & Fully connected autoencoder & CAE & Convolutional autoencoder \\
        GAE & Graph autoencoder & SC & Structural connectome \\
        MAE & Mean absoluter error & BMAE & Binary mean absoluter error\\
        PC & Pearson Correlation & NS & Nodal Strength \\
        CC & Closeness Centrality & CLC & Clustering Coefficient \\
        LE & Local Efficiency & EV & Eigenvalue \\
        FA & Fingerprinting accuracy & ID & Identifiability difference\\
        GP & Gender Prediction & AP & Age Prediction \\
        \bottomrule
    \end{tabular}
    \caption{Key abbreviations and their full forms.}
    \label{tab:abbreviations}
\end{table}

\section*{Ethical Approval}
The data used was obtained from the Human Connectome Project Young Adult dataset, with appropriate consent for both unrestricted and restricted data access.

\section*{Data and Code Availability}
The minimally preprocessed Magnetic Resonance Imaging data used in this study can be downloaded from the Human Connectome Project Young Adult Dataset:\\ \url{https://www.humanconnectome.org/study/hcp-young-adult}. Code used for additional preprocessing to generate structural connectomes is available upon request from the corresponding author. The code for different harmonization methods and their evaluation can be found in our Github repository:
\url{https://github.com/jagruti8/connectome_harmonization}.

\section*{Author Contributions}

Jagruti Patel: Conceptualization; Data curation; Formal analysis; Investigation; Methodology; Software; Visualization; Writing – original draft; Writing – review \& editing. 
Thomas A. W. Bolton: Conceptualization; Supervision; Validation; Writing – review \& editing.
Mikkel Schöttner: Conceptualization; Writing – review \& editing. 
Anjali Tarun: Data curation. 
Sebastien Tourbier: Data curation. 
Yasser Alemán-Gómez: Data curation. 
Jonas Richiardi: Methodology.
Patric Hagmann: Conceptualization; Methodology; Funding acquisition; Project administration; Resources; Supervision; Validation; Writing – review \& editing. 



\section*{Declaration of Competing Interests}

The authors declare having no conflict of interest.

\section*{Acknowledgements}

This work has been financially supported by the Swiss National Science Foundation grant number 197787.We thank Prof. Pascal Frossard and Mr. William Cappelletti of the Swiss Federal Technology Institute of Lausanne (EPFL) for their valuable discussions and insightful advice.

\newpage
\section*{Supplementary Material}

\renewcommand{\figurename}{Supplementary Figure}
\renewcommand{\thefigure}{\arabic{figure}}  
\setcounter{figure}{0}

\begin{figure}[!h]
\centerline{\includegraphics[width=0.95\textwidth]{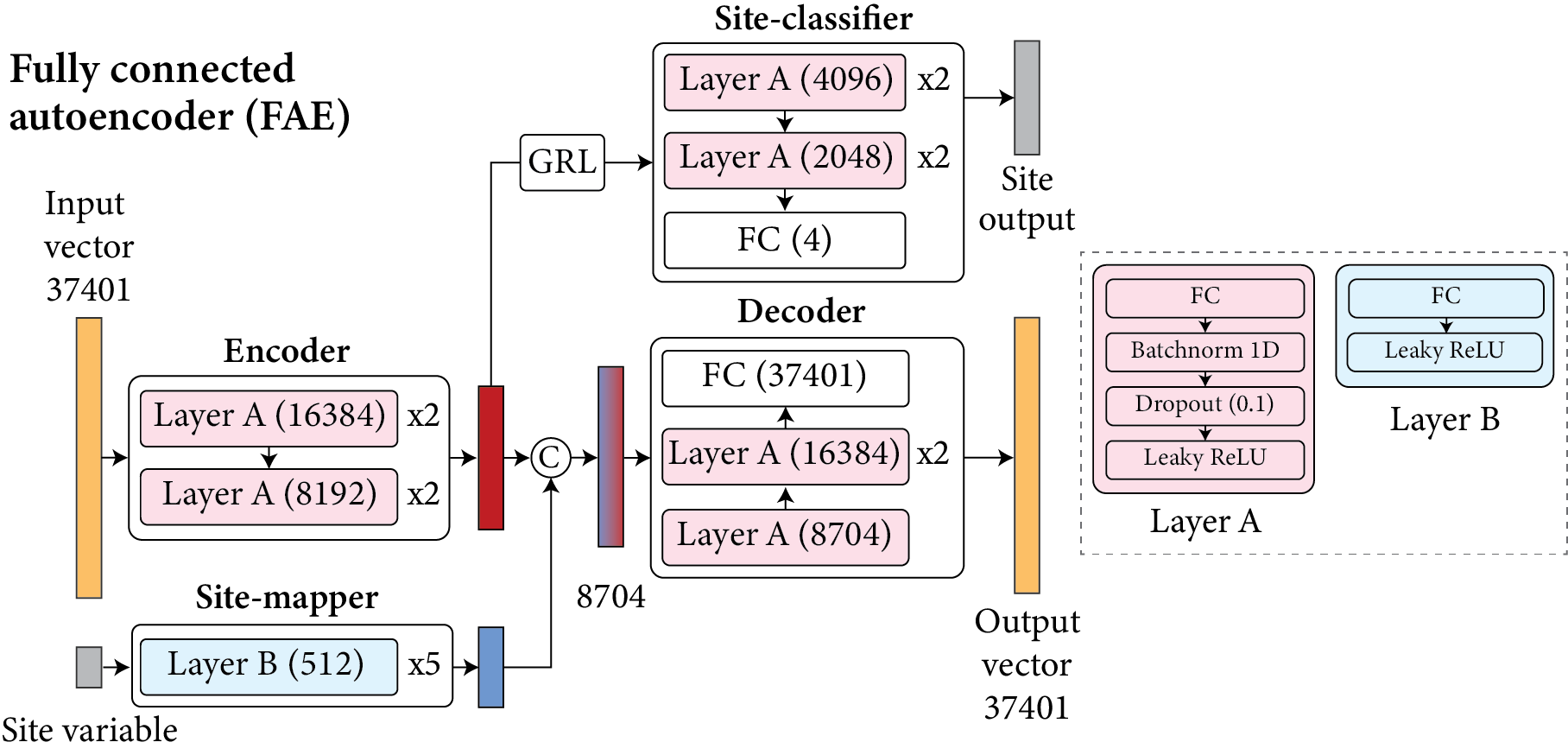}}
\caption{Fully connected autoencoder (FAE) architecture.}
\label{fig:FAE}
\end{figure}

\begin{figure}[!h]
\centerline{\includegraphics[width=0.95\textwidth]{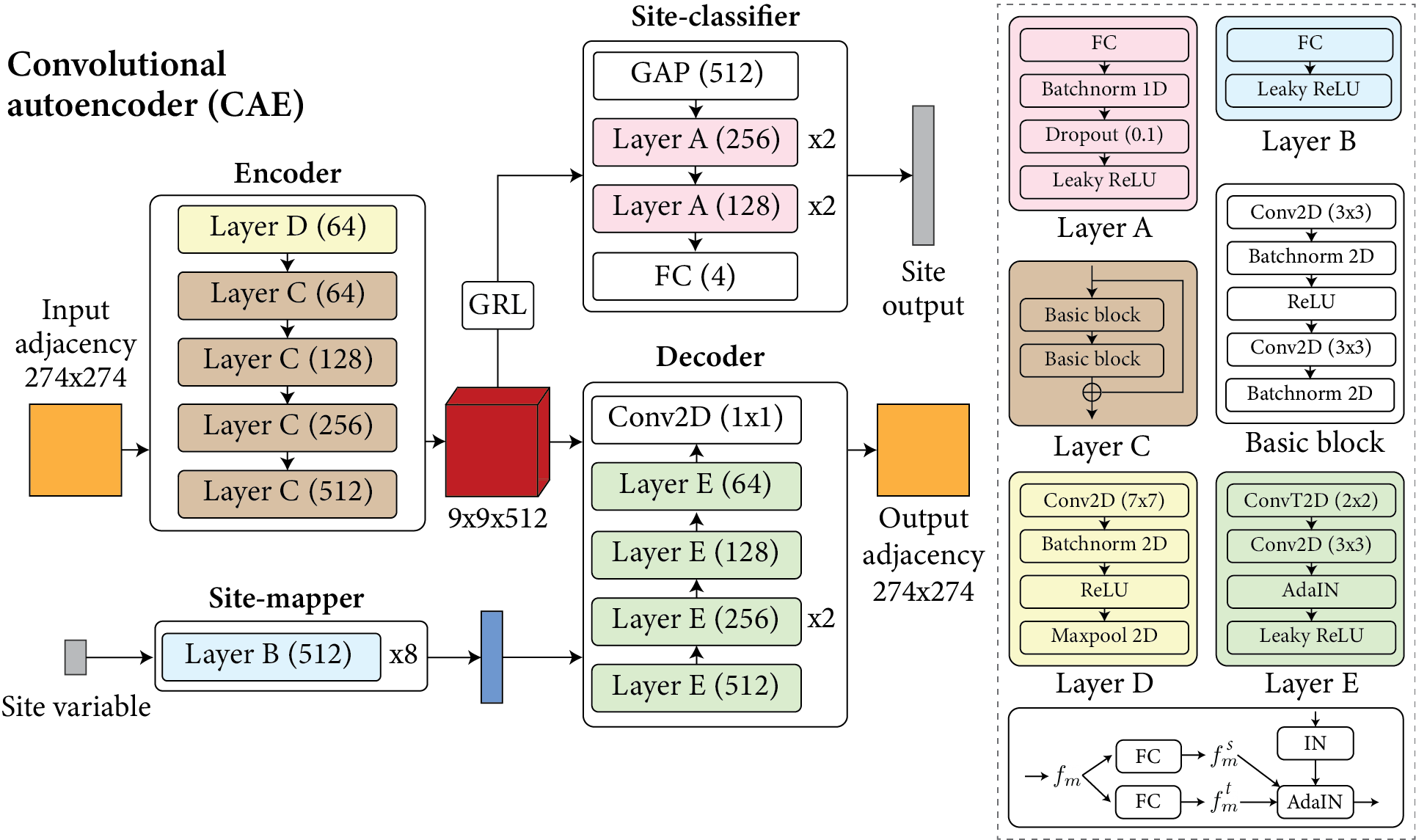}}
\caption{Convolutional autoencoder (CAE) architecture.}
\label{fig:CAE}
\end{figure}

\begin{figure}[!h]
\centerline{\includegraphics[width=0.95\textwidth]{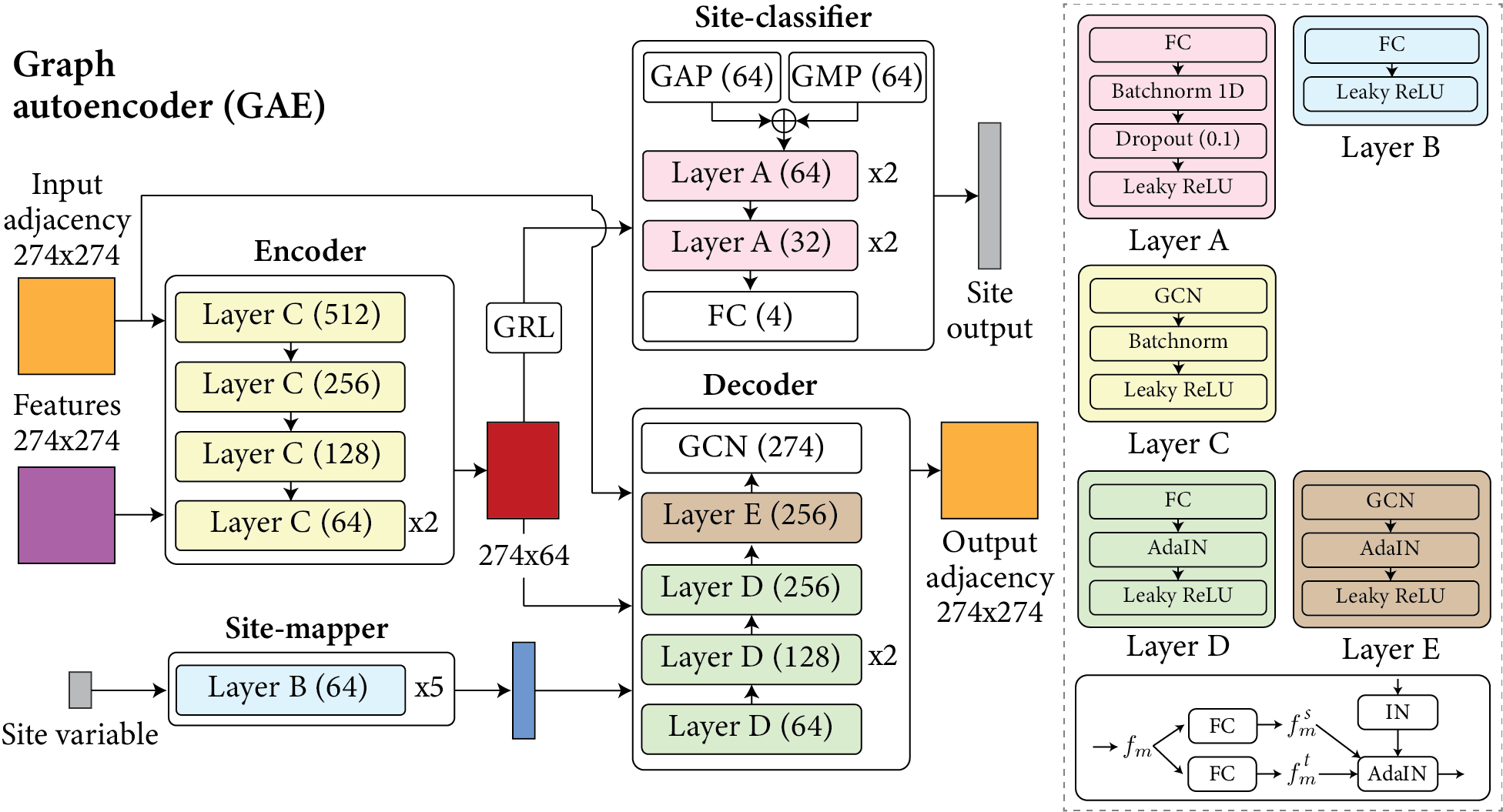}}
\caption{Graph autoencoder (GAE) architecture.}
\label{fig:GAE}
\end{figure}

\newpage
\printbibliography




\end{document}